\def\cvprPaperID{6248}
\def\confName{CVPR}
\def\confYear{2023}

\newif\ifreview 
\newif\ifarxiv \newcommand{\arxiv}{\arxivtrue}
\newif\ifcamera 
\newif\ifrebuttal 

\arxiv
\pdfoutput=1
\documentclass[10pt,twocolumn,letterpaper]{article}
\ifreview \usepackage[review]{cvpr} \fi
\ifarxiv \usepackage[pagenumbers]{cvpr} \fi
\ifrebuttal \usepackage[rebuttal]{cvpr} \fi
\ifcamera \usepackage{cvpr} \fi

\usepackage{graphicx}
\usepackage{amsmath}
\usepackage{amssymb}
\usepackage{booktabs}

\usepackage{times}
\usepackage{microtype}
\usepackage{epsfig}
\usepackage[table,xcdraw]{xcolor}
\usepackage{caption}
\usepackage{float}
\usepackage{placeins}
\usepackage{color, colortbl}
\usepackage{stfloats}
\usepackage{enumitem}
\usepackage{tabularx}
\usepackage{xstring}
\usepackage{multirow}
\usepackage{xspace}
\usepackage{url}
\usepackage{subcaption}
\usepackage{xcolor}
\usepackage{graphicx}
\usepackage[hang,flushmargin]{footmisc}

\ifcamera \usepackage[accsupp]{axessibility} \fi

% ...

\ifarxiv  \fi

\newcommand{\R}[1]{{%
    \textbf{%
        \ifstrequal{#1}{1}{\textcolor{red}{R#1}}{%
        \ifstrequal{#1}{2}{\textcolor{blue}{R#1}}{%
        \ifstrequal{#1}{3}{\textcolor{magenta}{R#1}}{%
        \ifstrequal{#1}{4}{\textcolor{teal}{R#1}}{%
                           \textcolor{cyan}{R#1}%
        }}}}%
    }%
}}

\usepackage{xr-hyper}

\makeatletter
\newcommand*{\addFileDependency}[1]{
  \typeout{(#1)}
  \@addtofilelist{#1}
  \IfFileExists{#1}{}{\typeout{No file #1.}}
}

\makeatother

\usepackage[pagebackref,breaklinks,colorlinks]{hyperref}
\usepackage[capitalize]{cleveref}
\crefname{section}{Sec.}{Secs.}
\crefname{table}{Table}{Tables}
\crefname{figure}{Fig.}{Figs.}

\begin{document}
\title{3D Concept Learning and Reasoning from Multi-View Images}

\author{Yining Hong\textsuperscript{1},  Chunru Lin\textsuperscript{2}, Yilun Du\textsuperscript{3}, \\ Zhenfang Chen\textsuperscript{5}, Joshua B. Tenenbaum\textsuperscript{3}, Chuang Gan\textsuperscript{4, 5}
 \\ 
\textsuperscript{1}UCLA, \textsuperscript{2,4}Shanghai Jiaotong University, \textsuperscript{3}MIT CSAIL,\\
\textsuperscript{4}UMass Amherst, \textsuperscript{5}MIT-IBM Watson AI Lab\\
\url{https://vis-www.cs.umass.edu/3d-clr/}
}

\twocolumn[{
\renewcommand\twocolumn[1][]{#1}
\maketitle
\begin{center}
    \includegraphics[width=\linewidth]{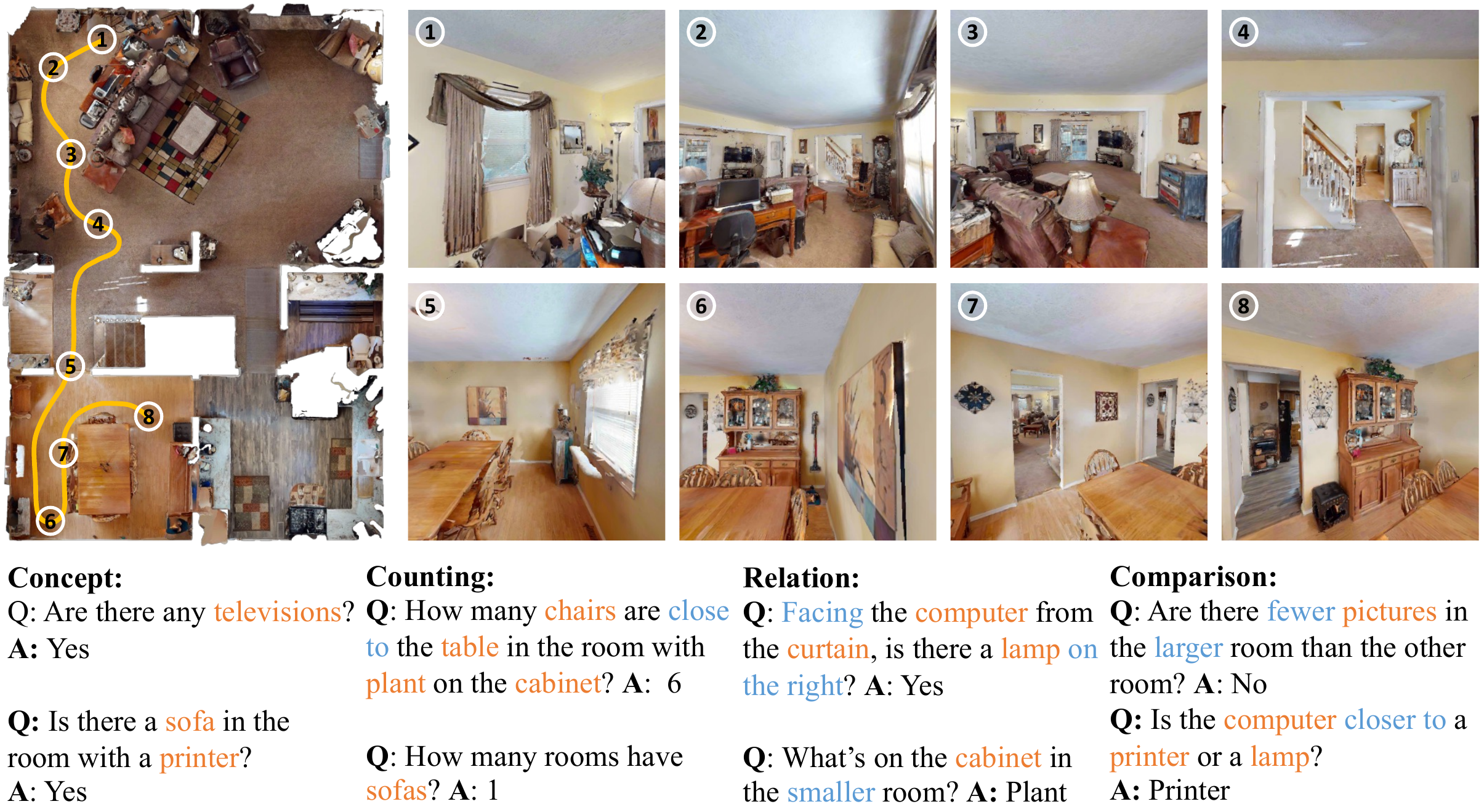}
    \vspace{-0.1in}
    \captionof{figure}{An exemplar scene with multi-view images and question-answer pairs of our 3DMV-VQA dataset. 3DMV-VQA contains four question types: concept, counting, relation, comparison. \textcolor{orange}{Orange} words denote semantic concepts; \textcolor{blue}{blue} words denote the relations.} \label{fig:teaser}
    \vspace{0.15in}
\end{center}
}]

\begin{abstract}

\end{abstract}
\vspace{-3mm}
Humans are able to accurately reason in 3D by gathering multi-view observations of the surrounding world.  Inspired by this insight, we introduce a new large-scale benchmark for 3D multi-view visual question answering (3DMV-VQA). This dataset is collected by an embodied agent actively moving and capturing RGB images in an environment using the Habitat simulator.  In total, it consists of approximately 5k scenes, 600k images, paired with 50k questions. We evaluate various state-of-the-art models for visual reasoning on our benchmark and find that they all perform poorly. We suggest that a principled approach for 3D reasoning from multi-view images should be to infer a compact 3D representation of the world from the multi-view images, which is further grounded on open-vocabulary semantic concepts, and then to execute reasoning on these 3D representations. As the first step towards this approach, we propose a novel 3D concept learning and reasoning (3D-CLR) framework that seamlessly combines these components via neural fields, 2D pre-trained vision-language models, and neural reasoning operators. Experimental results suggest that our framework outperforms baseline models by a large margin, but the challenge remains largely unsolved. We further perform an in-depth analysis of the challenges and highlight potential future directions.

\section{Introduction}
\label{sec:intro}

Visual reasoning, the ability to composite rules on internal representations to reason and answer questions about visual scenes, has been a long-standing challenge in the field of artificial intelligence and computer vision. Several datasets \cite{Zhu2016Visual7WGQ, Goyal2017MakingTV,Johnson2017CLEVRAD} have been proposed to tackle this challenge. However, they mainly focus on visual reasoning on 2D single-view images. Since 2D single-view images only cover a limited region of the whole space, such reasoning inevitably has several weaknesses, including occlusion, and failing to answer 3D-related questions about the entire scene that we are interested in. As shown in Fig. \ref{fig:teaser}, it's difficult, even for  humans, to count the number of chairs in a scene due to the object occlusion, and it's even harder to infer 3D relations like ``closer" from a single-view 2D image.

On the other hand, there's strong psychological evidence that human beings conduct visual reasoning in the underlying 3D representations \cite{spelke1993gestalt}.
 Recently, there have been several works focusing on 3D visual question answering \cite{Ye20213DQA, Azuma2022ScanQA3Q, yan2021clevr3d, etesam20223dvqa}. They mainly use traditional 3D representations (\textit{e.g.,} point clouds) for visual reasoning. This is inconsistent with the way human beings perform 3D reasoning in real life. Instead of being given an entire 3D representation of the scene at once, humans will actively walk around and explore the whole environment, ingesting image observations from different views and converting them into a holistic 3D representation that assists them in understanding and reasoning about the environment. Such abilities are crucial for many embodied AI applications, such as building assistive robots.
 

To this end, we propose the novel task of 3D visual reasoning from multi-view images taken by active exploration of an embodied agent.  Specifically, we generate a large-scale benchmark, 3DMV-VQA (3D multi-view visual question answering), that contains approximately 5k scenes and 50k question-answering pairs about these scenes. For each scene, we provide a collection of multi-view image observations. We generate this dataset by placing an embodied agent in the Habitat-Matterport environment \cite{Ramakrishnan2021HabitatMatterport3D}, which actively explores the environment and takes pictures from different views. We also obtain
scene graph annotations from the Habitat-Matterport 3D semantics dataset (HM3DSem) \cite{yadav2022habitat}, including ground-truth locations, segmentations, semantic information of the objects, as well as relationships among the objects in the environments, for model diagnosis. To evaluate the models' 3D reasoning abilities on the entire environment, we design several 3D-related question types, including concept, counting, relation and comparison.

Given this new  task, the key challenges we would like to investigate include: 1) how to efficiently obtain the compact visual representation to encode  crucial properties (\textit{e.g.,} semantics and relations) by integrating all incomplete observations of the environment in the process of active exploration for 3D visual reasoning? 2) How to ground the semantic concepts on these 3D representations that could be leveraged for downstream tasks, such as visual reasoning? 3) How to infer the relations among the objects, and perform step-by-step reasoning?

As the first step to tackling these challenges, we propose a novel model, 3D-CLR (3D Concept Learning and Reasoning). First, to efficiently obtain a compact 3D representation from multi-view images, we use a neural-field model based on compact voxel grids \cite{Sun2022DirectVG} which is both fast to train and effective at storing scene properties in its voxel grids. 
As for concept learning, we observe that previous works on 3D scene understanding \cite{Chen2020ScanRefer3O, Achlioptas2020ReferIt3DNL} lack the diversity and scale with regard to semantic concepts due to the limited amount of paired 3D-and-language data.  
Although large-scale vision-language models (VLMs) have achieved impressive performances for zero-shot semantic grounding on 2D images, leveraging these pretrained models for effective open-vocabulary 3D grounding of semantic concepts remains a challenge. To address these challenges, we propose to encode the features of a pre-trained 2D vision-language model (VLM) into the compact 3D representation defined across voxel locations.
Specifically, we use the CLIP-LSeg\cite{li2022language} model to obtain features on multi-view images, and propose an alignment loss to map the features in our 3D voxel grid to 2D pixels. By calculating the dot-product attention between the 3D per-point features and CLIP language embeddings, we can ground the semantic concepts in the 3D compact representation.
Finally, to answer the questions, we introduce a set of neural reasoning operators, including \textsc{filter}, \textsc{count}, \textsc{relation} operators and so on, which take the 3D representations of different objects as input and output the predictions. 

We conduct experiments on our proposed  3DMV-VQA benchmark. Experimental results 
 show that our proposed 3D-CLR outperforms all baseline models a lot. However, failure cases and model diagnosis show that challenges still exist concerning the grounding of small objects and the separation of close object instances. We provide an in-depth analysis of the challenges and discuss potential future directions.

 \noindent To sum up, we have the following contributions in this paper. 
 \vspace{-2mm}
\begin{itemize}[align=right,itemindent=0em,labelsep=2pt,labelwidth=1em,leftmargin=*,itemsep=0em] 
\item We propose the novel task of 3D concept learning and reasoning from multi-view images. 

\item By having robots actively explore the embodied environments, we collect a large-scale benchmark on 3D multi-view visual question answering (3DMV-VQA). 

\item We devise a model that incorporates a neural radiance field, 2D pretrained vision and language model, and neural reasoning operators to ground the concepts and perform 3D reasoning on the multi-view images. We  illustrate that our model outperforms all baseline models.

\item We perform an in-depth analysis of the challenges of this new task and highlight potential future directions. 
\end{itemize}
\section{Related Work}
\label{sec:related}

\noindent\textbf{Visual Reasoning}
There have been numerous tasks focusing on learning visual concepts from natural language, including visually-grounded question answering \cite{gan2017vqs,Ganju2017WhatsIA}, text-image retrieval \cite{Vendrov2016OrderEmbeddingsOI} and so on. Visual reasoning has drawn much attention recently as it requires human-like understanding of the visual scene. A wide variety of benchmarks have been created over the recent years~\cite{Johnson2017CLEVRAD, Goyal2017MakingTV,chencomphy,Zhu2016Visual7WGQ, Hong2021PTRAB,chen2020cops}. However, they mainly focus on visual reasoning from 2D single-view images, while there's strong psychological evidence that human beings perform visual reasoning on the underlying 3D representations. In this paper, we propose the novel task of visual reasoning from multi-view images, and collect a large-scale benchmark for this task. In recent years, numerous visual reasoning models have also been proposed, ranging from attention-based methods \cite{Hudson2018CompositionalAN,chen2019weakly}, graph-based methods \cite{Huang2020LocationAwareGC}, to models based on large pretrained vision-language model \cite{Li2022AlignAP,chen2023see}.
These methods model the reasoning process implicitly with neural networks. Neural-symbolic methods \cite{Yi2018NeuralSymbolicVD, Mao2019TheNC,chen2021grounding} explicitly perform symbolic reasoning on the objects representations and language representations. They use perception models to extract 2D masks as a first step, and then execute operators and ground concepts on these pre-segmented masks, but are limited to a set of pre-defined concepts on simple scenes. \cite{hong20223d} proposes to use the feature vectors from occupancy networks \cite{Mescheder2019OccupancyNL} to do visual reasoning in the 3D space. However, they also use a synthetic dataset, and learn a limited set of semantic concepts from scratch. We propose to learn 3D neural field features from 2D multi-view real-world images, and incorporate a 2D VLM
for open-vocabulary reasoning.

\noindent\textbf{3D Reasoning}
Understanding and reasoning about 3D scenes has been a long-standing challenge. Recent works focus on leveraging language to explore 3D scenes, such as object captioning \cite{Chen2020ScanRefer3O, Chen2021Scan2CapCD} and object localization from language \cite{Achlioptas2020ReferIt3DNL, Feng2021FreeformDG, Huang2021TextGuidedGN}. Our work is mostly related to 3D Visual Question Answering \cite{Ye20213DQA, Azuma2022ScanQA3Q, yan2021clevr3d, etesam20223dvqa} as we both focus on answering questions and reasoning about 3D scenes. However, these works use point clouds as 3D representations, which diverts from the way human beings perform 3D reasoning. Instead of being given an entire 3D representation all at once, human beings would actively move and explore the environment, integrating multi-view information to get a compact 3D representation. Therefore, we propose 3D reasoning from multi-view images. In addition, since 3D assets paired with natural language descriptions are hard to get in real-life scenarios, previous works struggle to ground open-vocabulary concepts. In our work, we leverage 2D VLMs for zero-shot open-vocabulary concept grounding in the 3D space.


\noindent\textbf{Embodied Reasoning} Our work is also closely related to Embodied Question Answering (EQA)~\cite{Das2018EmbodiedQA,yu2019multi} and Interactive Question Answering (IQA)~\cite{gordon2018iqa,konstantinova2013interactive}, which also involve an embodied agent exploring the environment and answering the question. However, the reasoning mainly focuses on the outcome or the history of the navigation on 2D images and does not require a holistic 3D understanding of the environment. There are also works~\cite{suglia2021embodied,song2022one,shridhar2020alfred,zheng2022vlmbench,gao2022dialfred,dingembodied} targeting instruction following in embodied environments, in which an agent is asked to perform a series of tasks based on language instructions. Different from their settings, for our benchmark an embodied agent actively explores the environment and takes multi-view images for 3D-related reasoning.


\noindent\textbf{Neural Fields}
Our approach utilizes neural fields to parameterize an underlying 3D compact representations of scenes for reasoning.   Neural field models (\textit{e.g.,} \cite{mildenhall2020nerf}) have gained much popularity since they can reconstruct a volumetric 3D scene representation from a set of images. Recent works \cite{Garbin2021FastNeRFHN, Hedman2021BakingNR, Yu2021PlenOctreesFR, Sun2022DirectVG} have pushed it further by using classic voxel-grids to explicitly store the scene properties (\textit{e.g.}, density, color and feature) for rendering, which allows for real-time rendering and is utilized by this paper.
Neural fields have also been used to represent dynamic scenes \cite{niemeyer2019occupancy,du2021nerflow}, appearance \cite{sitzmann2019srns,mildenhall2020nerf,Niemeyer2020DVR,yariv2020multiview,saito2019pifu}, physics \cite{Kollmannsberger2021PhysicsInformedNN}, robotics \cite{Jiang2021SynergiesBA, simeonov2021neural}, acoustics \cite{luo2022learning} and more general multi-modal signals \cite{du2021gem}. There are also some works that integrate semantics or language in neural fields \cite{Jain2022ZeroShotTO, Wang2021CLIPNeRFTD}. However, they mainly focus on using language for manipulation, editing or generation. 
\cite{hong20223d} leverages neural descriptor field \cite{simeonov2021neural} for 3D concept grounding. However, they require ground-truth occupancy values to train the neural field, which can not be applied to real-world scenes. In this paper, we propose to leverage voxel-based neural radiance field \cite{Sun2022DirectVG} to get the compact representations for 3D visual reasoning. 

\section{Dataset Generation}
\label{sec:dataset}

\subsection{Multi-View Images}

Our dataset includes 5k 3D scenes from the Habitat-Matterport 3D  Dataset (HM3D) dataset \cite{Ramakrishnan2021HabitatMatterport3D}, and approximately 600k images rendered from the 3D scenes. The images are rendered via Habitat \cite{szot2021habitat, habitat19iccv}. 

\noindent\textbf{Scene Generation} 
We build our benchmark on top of the HM3DSem dataset\cite{yadav2022habitat}, which is a large-scale dataset of 3D real-world indoor scenes with densely annotated semantics. It consists of 142,646 object instance annotations across 216 3D spaces and 3,100 rooms within those spaces. HM3D dataset uses texture information to annotate pixel-accurate object boundaries, which provides large-scale object annotations and ensures the scale, quality, and diversity of 3D visual reasoning questions of our benchmark.

To construct a benchmark that covers questions of different difficulty levels, it's crucial that we include 3D scenes of different scales in our benchmark. We start with single rooms in HM3D scenes, which has an appropriate amount of semantic concepts and relationships to base some simple questions on.
To get the scale of single rooms, we calculate bounding boxes of rooms according to floor instance segmentations. We then proceed to generate bounding boxes for scenes with multiple adjacent rooms. For more complex holistic scene understanding, we also include whole-house scenes, which may contain tens of rooms. Overall, the 3DMV-VQA benchmark contains three levels of scenes (2000 single-room scenes, 2000 multi-room scenes and 100 whole-house scenes).


\noindent\textbf{Image Rendering}
After we get the bounding box of each scene, we load the scene into the Habitat simulator. We also put a robot agent with an RGB sensor at a random initial point in the bounding box. The data is collected via exploration of the robot agent. Specifically, at each step of the data collection process, we sample a navigable point and make the agent move to the point along the shortest path. When the agent has arrived at a point, we rotate the agent $30^{\circ} $ along z-axis for 12 times so that the agent can observe the $360^{\circ}$ view of the scene at the position. It can also look up and down, with a random mild angle from [$-10^{\circ}$,$10^{\circ}$] along the x-axis. A picture is taken each time the agent rotates to a new orientation. In total 12 pictures are taken from each point. While traveling between points, the robot agent further takes pictures. We also exploit a policy such that when the camera is too far from or too close to an object and thus the agent cannot see anything, we discard the bad-view images.  


\subsection{Questions and Answers}
 We pair each scene with machine-generated questions from pre-defined templates. All questions are open-ended and can be answered with a single word (samples in Fig. \ref{fig:teaser}).
 
\noindent\textbf{Concepts and Relationships} To generate questions and answers, we utilize the semantic annotations of HM3DSem\cite{yadav2022habitat} to get the semantic concepts and their bounding boxes, as well as the bounding boxes of the rooms.  We merge semantic concepts with similar meanings (\textit{e.g.,}, L-shaped sofa to sofa, desk chair / computer chair {e.g.} to chair).  We also define 11 relationships: inside, above, below, on the top of, close, far, large, small, between, on the left, and on the right. Before generating questions, we first generate a scene graph for each scene containing all concepts and relationships.

\noindent\textbf{Question Types}
We define four types of questions: concept, counting, relation and comparison.
\vspace{-2mm}
\begin{itemize}
[align=right,itemindent=0em,labelsep=2pt,labelwidth=1em,leftmargin=*,itemsep=0em] 
\item \textbf{Concept.} Conceptual questions query whether there's an object of a certain semantic concept in the scene, or whether there's a room containing the objects of the semantic concept.
\item \textbf{Counting.} Counting-related questions ask about how many instances of a semantic concept are in the scene, or how many rooms contain objects of the semantic concept.
\item \textbf{Relation.} Relational questions ask about the 11 relationships and their compositions. Based on the number of relations in a question, we have one-hop to three-hop questions for the relation type.
\item \textbf{Comparison.} The comparison question type focuses on the comparison of two objects, two semantic concepts or two rooms. It can be combined with the relational concepts to compare two objects (\textit{e.g.,} larger, closer to, more left \textit{etc}). It also compares the number of instances of two semantic concepts, or the number of objects of certain concepts in different rooms.
\end{itemize}

\noindent\textbf{Bias Control.} Similar to previous visual reasoning benchmarks \cite{Johnson2017CLEVRAD, hong20223d}, we use machine-generated questions since the generation process is fully controllable so that we can avoid dataset bias. Questions are generated from pre-defined templates, and transformed into natural
language questions with associated semantic concepts and relationships from the scene.    We manually define 41 templates for question generation. We use depth-first search to generate questions. We perform bias control based on three perspectives: template counts, answer counts, and concept counts. For selecting templates, we sort the templates each time we generate a question to ensure a balanced question distribution. We force a flat
answer distribution for each template by rejection sampling. Specifically, once we generate a question and an answer, if the number of the questions having the same answer and template is significantly larger than other answers, we discard it and continue searching. Once we find an answer that fits in the ideal answer distribution, we stop the depth-first searching for this question. We also force a flat concept distribution for each template using the same method. In addition to controlling the number of concepts mentioned in the templates, we also control the number of relation tuples consisting of the same concept sets. 
\section{Method}

\begin{figure*}[t]
\centering
\includegraphics[width=\linewidth]{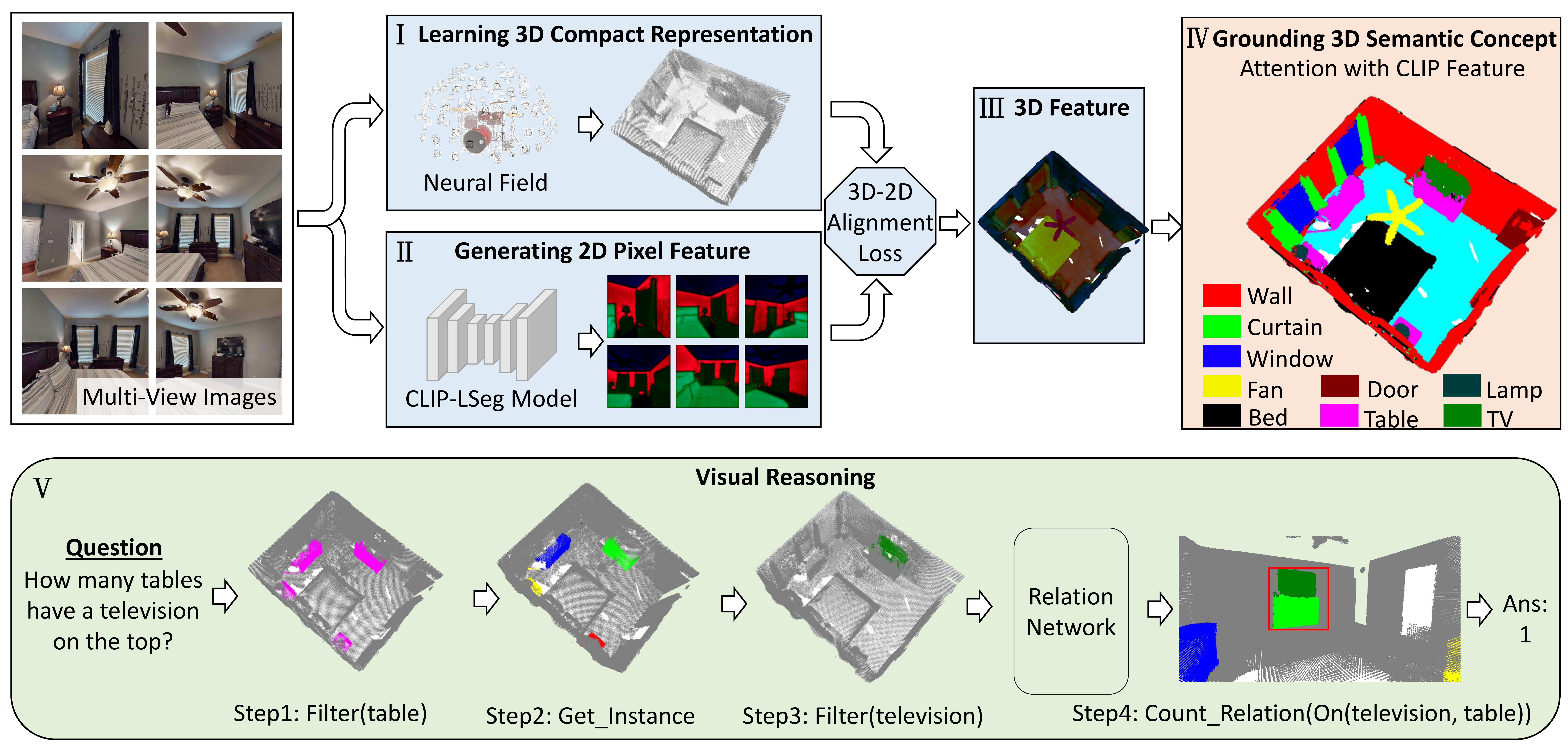}
\vspace{-5mm}
\caption{An overview of our 3D-CLR framework. First, we learn a 3D compact scene representation from multi-view images using neural fields (I). Second, we use CLIP-LSeg model to get per-pixel 2D features (II). We utilize a 3D-2D alignment loss to assign features to the 3D compact representation (III). By calculating the dot-product attention between the 3D per-point features and CLIP language embeddings, we could get the concept grounding in 3D (IV). Finally, the reasoning process is performed via a set of neural reasoning operators, such as \textsc{Filter}, \textsc{Get\_Instance} and \textsc{Count\_Relation} (V). Relation operators are learned via relation networks.}
\vspace{-5mm}
\label{fig:framework}
\end{figure*}

Fig.~\ref{fig:framework} illustrates an overview of our framework. Specifically, our framework consists of three steps.  First, we learn a 3D compact representation from multi-view images using neural field. And then we propose to leverage pre-trained 2D vision-and-language model to ground concepts on 3D space. This is achieved by 1) generating 2D pixel features using CLIP-LSeg; 2) aligning the features of 3D voxel grid and 2D pixel features from CLIP- LSeg~\cite{li2022language}; 3) dot-product attention between the 3D features and CLIP language features~\cite{li2022language}. Finally, to perform visual reasoning, we propose neural reasoning operators, which execute the question step by step on the 3D compact representation and outputs a final answer. For example, we use \textsc{Filter} operators to ground semantic concepts on the 3D representation, \textsc{Get\_Instance} to get all instances of a semantic class, and \textsc{Count\_Relation} to count how many pairs of the two semantic classes have the queried relation.

\subsection{Learning 3D Compact Scene Representations}


Neural radiance fields  \cite{mildenhall2020nerf} are capable of learning a 3D representation that can reconstruct a volumetric 3D scene representation from a set of images. Voxel-based methods \cite{Garbin2021FastNeRFHN, Hedman2021BakingNR, Yu2021PlenOctreesFR, Sun2022DirectVG} speed up the learning process by explicitly storing the scene properties (\textit{e.g.}, density, color and feature) in its voxel grids. We leverage Direct Voxel Grid Optimization (DVGO) \cite{Sun2022DirectVG} as our backbone for 3D compact representation for its fast speed. DVGO stores the learned density and color properties in its grid cells. The rendering of multi-view images is by interpolating through the voxel grids to get the density and color for each sampled point along each sampled ray, and integrating the colors based on the rendering alpha weights calculated from densities according to quadrature rule \cite{Max1995OpticalMF}. The model is trained by minimizing the L2 loss between the rendered multi-view images and the ground-truth multi-view images. By extracting the density voxel grid, we can get the 3D compact representation (\textit{e.g.,} By visualizing points with density greater than 0.5, we can get the 3D representation as shown in Fig. \ref{fig:framework} I. ) 

\subsection{3D Semantic Concept Grounding}
Once we extract the 3D compact representation of the scene, we need to ground the semantic concepts for reasoning from language. 
Recent work from \cite{hong20223d} has proposed to ground concepts from paired 3D assets and question-answers. Though promising results have been achieved on synthetic data, it is not feasible for open-vocabulary 3D reasoning in real-world data, since it is hard to collect large-scale 3D vision-and-language paired data.  To address this challenge, our idea is to leverage  pre-trained 2D vision and language model \cite{Radford2021LearningTV, Ramesh2021ZeroShotTG} for 3D concept grounding in real-world scenes.  But how can we map 2D concepts into 3D neural field representations? Note that 3D compact representations can be learned from 2D multi-view images and that each 2D pixel actually corresponds to several 3D points along the ray. Therefore, it's possible to get 3D features from 2D per-pixel features. Inspired by this, we first add a feature voxel grid representation to DVGO, in addition to density and color, to represent 3D features. 
 We then apply CLIP-LSeg\cite{li2022language} to learn per-pixel 2D features, which can be attended to by CLIP concept embeddings. We use an alignment loss to align 3D features with 2D features so that we can perform concept grounding on the 3D representations.

\noindent\textbf{2D Feature Extraction.}
To get per-pixel features that can be attended by concept embeddings, we use the features from language-driven semantic segmentation (CLIP-LSeg) \cite{li2022language}, which learns 2D per-pixel features from a pre-trained vision-language model (\textit{i.e.,} \cite{Radford2021LearningTV}). Specifically, it
uses the text encoder from CLIP, trains an image encoder to produce an embedding vector for each pixel, and calculates the scores of word-pixel correlation by dot-product. By outputting the semantic class with the maximum score of each pixel, CLIP-LSeg is able to perform zero-shot 2D semantic segmentation.

\noindent\textbf{3D-2D Alignment.}
In addition to density and color, we also store a 512-dim feature in each grid cell in the compact representation. To align the 3D per-point features with 2D per-pixel features, we calculate an L1 loss between each pixel and each 3D point sampled on the ray of the pixel. The overall L1 loss along a ray is the weighted sum of all the pixel-point alignment losses, with weights same as the rendering weights: $\mathcal{L}_{\text {feature}}=\sum_{i=1}^K w_i(\|\boldsymbol{f_i}-F(\boldsymbol{r})\|),$
where $\boldsymbol{r}$ is a ray corresponding to a 2D pixel, $F(\boldsymbol{r})$ is the 2D feature from CLIP-LSeg, $K$ is the total number of sampled points along the ray and $\boldsymbol{f_i}$ is the feature of point $i$ by interpolating through the feature voxel grid, $w_i$ is the rendering weight.

\noindent\textbf{Concept Grounding through Attention.}  Since our feature voxel grid representation is learnt from CLIP-LSeg, by calculating the dot-product attention $<\boldsymbol{f}, \boldsymbol{v}> $ between per-point 3D feature $\boldsymbol{f}$ and the CLIP concept embeddings $\boldsymbol{v}$, we can get zero-shot view-independent concept grounding and semantic segmentations in the 3D representation, as is presented in Fig. \ref{fig:framework} IV. 

\subsection{Neural Reasoning Operators}
Finally, we use the grounded semantic concepts for 3D reasoning from language. We first transform questions into a sequence of operators that can be executed on the 3D representation for reasoning. We adopt a LSTM-based semantic parser   \cite{Yi2018NeuralSymbolicVD} for that. As \cite{Mao2019TheNC, hong20223d}, we further devise a set of operators which can be executed on the 3D representation.  Please refer to \textbf{Appendix} for a full list of operators.

\noindent\textbf{Filter Operators.}  We filter all the grid cells with a certain semantic concept.

\noindent\textbf{Get\_Instance Operators.} We implement this by utilizing DBSCAN \cite{Ester1996ADA}, an unsupervised algorithm which assigns clusters to a set of points. Specifically, given a set of points in the 3D space, it can group together the points that are closely packed together for instance segmentation.

\noindent\textbf{Relation Operators.} We cannot directly execute the relation on the 3D representation as we have not grounded relations. Thus, we represent each relation using a distinct neural module (which is practical as the vocabulary of relations is limited \cite{Landau1993WhatA}). We first concatenate the voxel grid representations of all the referred objects and feed them into the relation network.
The relation network consists of three 3D convolutional layers and then three 3D deconvolutional layers. A score is output by the relation network indicating whether the objects have the relationship or not. Since vanilla 3D CNNs are very slow, we use Sparse Convolution \cite{spconv2022} instead. Based on the relations asked in the questions, different relation modules are chosen.

\section{Experiments}
\label{sec:exp}

 \subsection{Experimental Setup}
\noindent\textbf{Evaluation Metric.} We report the visual question answering accuracy on the proposed 3DMV-VQA dataset
w.r.t the four types of questions. The train/val/test split is 7:1:2. 

\noindent\textbf{Implementation Details} For 3D compact representations, we adopt the same architectures as DVGO, except skipping the coarse reconstruction phase and directly training the fine reconstruction phase. After that, we freeze the density voxel grid and color voxel grid, for the optimization of the feature voxel grid only. The feature grid has a world size of 100 and feature dim of 512. We train the compact representations for 100,000 iterations and the 3D features for another 20,000 iterations. For LSeg, we use the official demo model, which has the ViT-L/16 image encoder and CLIP’s ViT-B/32 text encoder. We follow the official script for inference and use multi-scale inference. For DBSCAN, we use an epsilon value of 1.5, minimum samples of 2, and we use L1 as the clustering method. For the relation networks, each relation is encoded into a three-layer sparse 3D convolution network with hidden size 64. The output is then fed into a one-layer linear network to produce a score, which is normalized by sigmoid function. We use cross-entropy loss to train the relation networks, and we use the one-hop relational questions with ``yes/no" answers to train the relation networks.

\subsection{Baselines}
Our baselines range from vanilla neural networks, attention-based methods, fine-tuned from large-scale VLM, and graph-based methods, to neural-symbolic methods.
\begin{itemize}
[align=right,itemindent=0em,labelsep=2pt,labelwidth=1em,leftmargin=*,itemsep=0em]
\item \textbf{LSTM}. The question is transferred to word embeddings which are input into a word-level LSTM \cite{Hochreiter1997LongSM}. The last LSTM hidden state is fed into a multi-layer perceptron (MLP) that outputs a distribution
over answers. This method is able to model question-conditional bias since it uses no image information.
\item \textbf{CNN+LSTM}. The question is encoded by the final hidden states from LSTM. We use a resnet-50 to extract frame-level features of images and average them over the time dimension. The
features are fed to an MLP to predict the final answer. This is a simple baseline that
examines how vanilla neural networks perform on 3DMV-VQA.
\item \textbf{3D-Feature+LSTM}. We use the 3D features we get from 3D-2D alignment and downsample the voxel grids using 3D-CNN as input, concatenated with language features from LSTM and fed to an MLP.
\item  \textbf{MAC} \cite{Hudson2018CompositionalAN}. MAC utilizes a Memory, Attention and Composition cell to perform iterative reasoning process. Like CNN+LSTM, we use the average pooling over multi-view images as the feature map. 

\item \textbf{MAC(V)}. We treat the multi-view images along a trajectory as a video. We modify the MAC model by applying a temporal attention unit across the video frames to generate a latent encoding for the video.
\item \textbf{NS-VQA}\cite{Yi2018NeuralSymbolicVD}. This is a 2D version of our 3D-CLR model. We use CLIP-LSeg to ground 2D semantic concepts from multi-view images, and the relation network also takes the 2D features as input. We execute the operators on each image and max pool from the answers to get our final predictions.
\item \textbf{ALPRO} \cite{Li2022AlignAP}. ALPRO is a video-and-language pre-training framework. A transformer model is pretrained  on large webly-source video-text pairs and can be used for downstream tasks like Video Question answering.
\item \textbf{LGCN} \cite{Huang2020LocationAwareGC}. LGCN represents the contents in the video as a location-aware graph by incorporating the location information of an
object into the graph construction.
\end{itemize}

\subsection{Experimental Results}

\begin{table*}[t]
	\begin{center}\small
 
	\begin{tabular}{lccccc}
	\toprule
      Methods   & Concept & Counting & Relation& Comparison & Overall\\ 
     
    \midrule
        Q-type (rand.)  &49.4 &10.7 &21.6 & 49.2 &26.4\\ 
        Q-type (freq.)  &50.8 &11.3  &23.9 &50.3 &28.2\\
        LSTM           &53.4&15.3&24.0&55.2 &29.8\\
        \midrule
        CNN+LSTM  &57.8&22.1&35.2&59.7 &37.8\\ 
        MAC &62.4&19.7&47.8&62.3 &46.7\\    
        MAC(V) &60.0&24.6&51.6&65.9 &50.0\\
        NS-VQA &59.8&21.5&33.4&61.6 &38.0\\ 
        ALPRO &65.8 &12.7&42.2&68.2 &43.3\\
        LGCN &56.2&19.5&35.5&66.7 &39.1\\
        3D-Feature+LSTM  &61.2 &22.4 & 49.9 & 61.3 &48.2\\ 
        \midrule
        3D-CLR (Ours)  & \textbf{66.1} & \textbf{41.3} &\textbf{57.6}& \textbf{72.3} &\textbf{57.7}\\ 

    \bottomrule
	\end{tabular}
	\end{center}
	\vspace{-13pt}
	\caption{Question-answering accuracy of 3D visual reasoning baselines on different question types.}
	\vspace{-12pt}
	\label{tab:reasoning}
\end{table*}

\noindent\textbf{Result Analysis.} We summarize the performances for each question type of baseline models in Table \ref{tab:reasoning}. All models are trained on the training set until convergence, tuned on the validation set, and evaluated on the test set. We provide detailed analysis below.

First, for the examination of language-bias of the dataset, we find that the performance of LSTM is only slightly higher than random and frequency, and all other baselines outperform LSTM a lot. This suggests that there's little language bias in our dataset. Second, we observe that encoding temporal information in MAC (\textit{i.e.,} MAC(V)) is better than average-pooling of the features, especially in counting and relation. This suggests that average-pooling of the features may cause the model to lose information from multi-view images, while attention on multi-view images helps boost the 3D reasoning performances. Third, we also find that fine-tuning on large-scale pretrained model (\textit{i.e.,} ALPRO) has relatively high accuracies in concept-related questions, but for counting it's only slightly higher than the random baseline, suggesting that pretraining on large-scale video-language dataset may improve the model's perception ability, but does not provide the model with the ability to tackle with more difficult reasoning types such as counting. Next, we find that LGCN has poor performances on the relational questions, indicating that building a location-aware graph over 2D objects still doesn't equip the model with 3D location reasoning abilities. Last but not least, we find that 3D-based baselines are better than their 2D counterparts. 3D-Feature+LSTM performs well on the 3D-related questions, such as counting and relation, than most of the image-based baselines. Compared with 3D-CLR, NS-VQA can perform well in the conceptual questions. However, it underperforms 3D-CLR a lot in counting and relation, suggesting that these two types of questions require the holistic 3D understanding of the entire 3D scenes. Our 3D-CLR outperforms other baselines by a large margin, but is still far from satisfying. From the accuracy of the conceptual question, we can see that it can only ground approximately 66\% of the semantic concepts. This indicates that our 3DMV-VQA dataset is indeed very challenging.

\begin{figure*}[t]
\centering
\includegraphics[width=\linewidth]{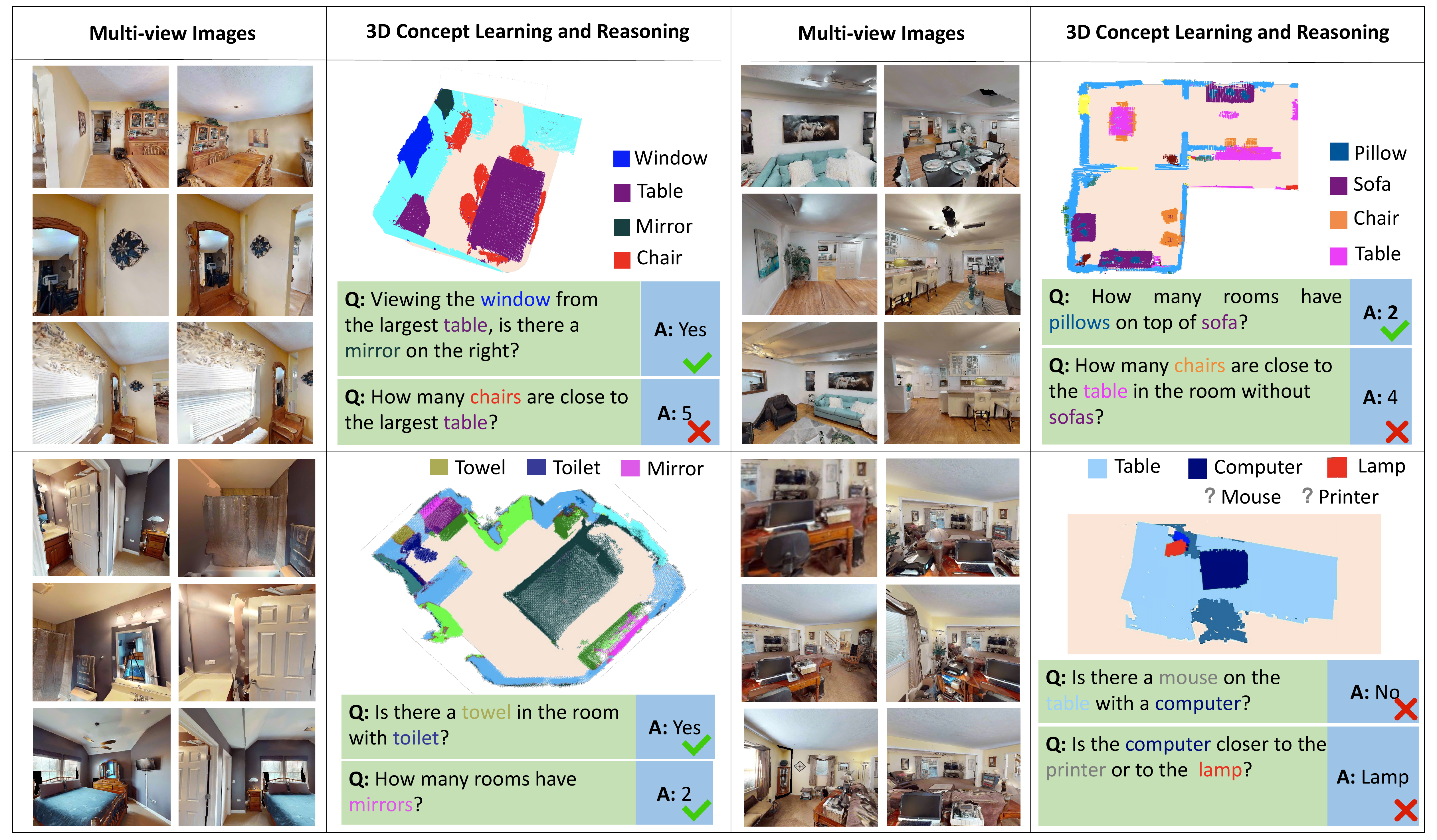}
\vspace{-6mm}
\caption{Qualitative examples of our 3D-CLR. We can see that 3D-CLR can ground most of the concepts and answer most questions correctly. However, it still fails sometimes, mainly because it cannot separate close object instances and ground small objects. }
\vspace{-4mm}
\label{fig:qualitative}
\end{figure*}

\noindent\textbf{Qualitative Examples.} In Fig. \ref{fig:qualitative}, we show four qualitative examples. From the examples, we show that our 3D-CLR can infer an accurate 3D representation from multi-view images, as well as ground semantic concepts on the 3D representations to get the semantic segmentations of the entire scene.  Our 3D-CLR can also learn 3D relationships such as ``close", ``largest", ``on top of" and so on. However, 3D-CLR also fails on some questions. For the third scene in the qualitative examples, it fails to ground the concepts ``mouse" and ``printer". Also, it cannot accurately count the instances sometimes. We give detailed discussions below. 


\subsection{Discussions}
\begin{figure}[t]
\centering
\includegraphics[width=0.98\linewidth]{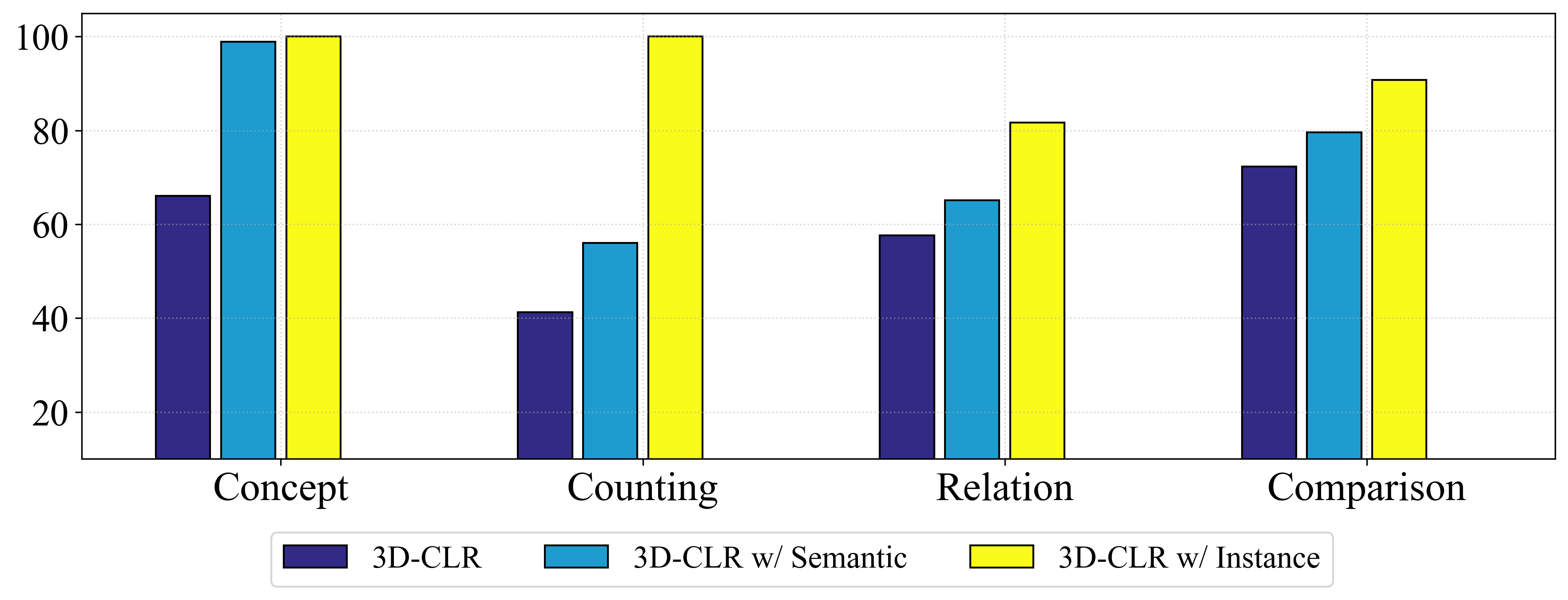}
\vspace{-3mm}

\caption{Model diagnosis of our 3D-CLR. }
\vspace{-5mm}

\label{fig:ablative}
\end{figure}

We perform an in-depth analysis to understand the challenge of this dataset. We leverage the modular design of our 3D-CLR, replacing individual components of the framework with ground-truth annotations for model diagnosis. The result is shown in Fig \ref{fig:ablative}. 3D-CLR w/ Semantic denotes our model with ground-truth semantic concepts from HM3DSem annotations. 3D-CLR w/ Instance denotes that 
we have ground-truth instance segmentations of semantic concepts. From Fig.~\ref{fig:qualitative} and Fig.~\ref{fig:ablative}, we summarize several key challenges of our benchmark:

\noindent \textbf{Very close object instances} From Fig.~\ref{fig:ablative}, we can see that even with ground-truth semantic labeling of the 3D points, 3D-CLR still has unsatisfying results on counting questions. This suggests that the instance segmentations provided by DBSCAN are not accurate enough. From the top two qualitative examples in Fig.~\ref{fig:qualitative}, we can also see that if two chairs contact each other, DBSCAN will not tell them apart and thus have poor performance on counting. One crucial future direction is to improve unsupervised instance segmentations on very close object instances.

\noindent \textbf{Grounding small objects}
Fig.~\ref{fig:ablative} suggests that 3D-CLR fails to ground a large portion of the semantic concepts, which hinders the performance. From the last example in Fig. ~\ref{fig:qualitative}, we can see that 3D-CLR fails to ground small objects like ``computer mouse". Further examination indicates there are two possible reasons: 1) CLIP-LSeg fails to assign the right features to objects with limited pixels; 2) The resolution of feature voxel grid is not high enough and therefore small objects cannot be represented in the compact representation. An interesting future direction would be learning exploration policies that enable the agents to get closer to uncertain objects that cannot be grounded.

\noindent \textbf{Ambiguity on 3D relations} 
Even with ground-truth semantic and instance segmentations, the performance of the relation network still needs to be improved. We find that most of the failure cases are correlated to the ``inside" relation. From the segmentations in Fig.~\ref{fig:qualitative}, we can see that 3D-CLR is unable to ground the objects in the cabinets. A potential solution can be joint depth and segmentation predictions. 




\section{Conclusion}
\label{sec:conclusion}
In this paper, we introduce the novel task of 3D reasoning from multi-view images. By placing embodied robot that actively explores indoor environments, we collect a large-scale benchmark named 3DMV-VQA. We also propose a new 3D-CLR model that incorporates neural field, 2D VLM, as well as reasoning operators for this task and illustrate its effectiveness. Finally, we perform an in-depth analysis to understand the challenges of this dataset and also point out potential future directions. We hope that 3DMV-VQA can be used to push the frontiers of 3D reasoning.

\paragraph{Acknowledgements.} This work was supported by the MIT-IBM Watson AI Lab, DARPA MCS, DSO grant DSOCO21072, and gift funding from MERL, Cisco, Sony, and Amazon. We would also like to thank the computation support from AiMOS, a server cluster for the IBM Research AI Hardware Center.

{\small
\bibliographystyle{ieee_fullname}
\bibliography{11_references}
}

\ifarxiv \clearpage 








\onecolumn
\appendix

\begin{center}
	{
		\Large{\textbf{Supplementary Material for \\``3D Concept Learning and Reasoning from Multi-View Images"}}
	}
\end{center}
    
    



{
    \hypersetup{linkcolor=black}
    \tableofcontents
}
\clearpage



\section{Dataset}
\label{sec:dataset}
\subsection{Dataset Statistics}
Figure \ref{fig:statistics}, we show some statistics about our dataset. From Figure \ref{fig:statistics}(a), we can see that relation type takes up the most portion of the questions, which is reasonable since our dataset focuses on 3D reasoning, and spatial relation is a crucial perspective. From Figure \ref{fig:statistics} (b), we can see that our questions cover a wide range of word lengths.
\begin{figure}[htbp]
\centering
\begin{subfigure}{.35\textwidth}
\includegraphics[width=\linewidth]{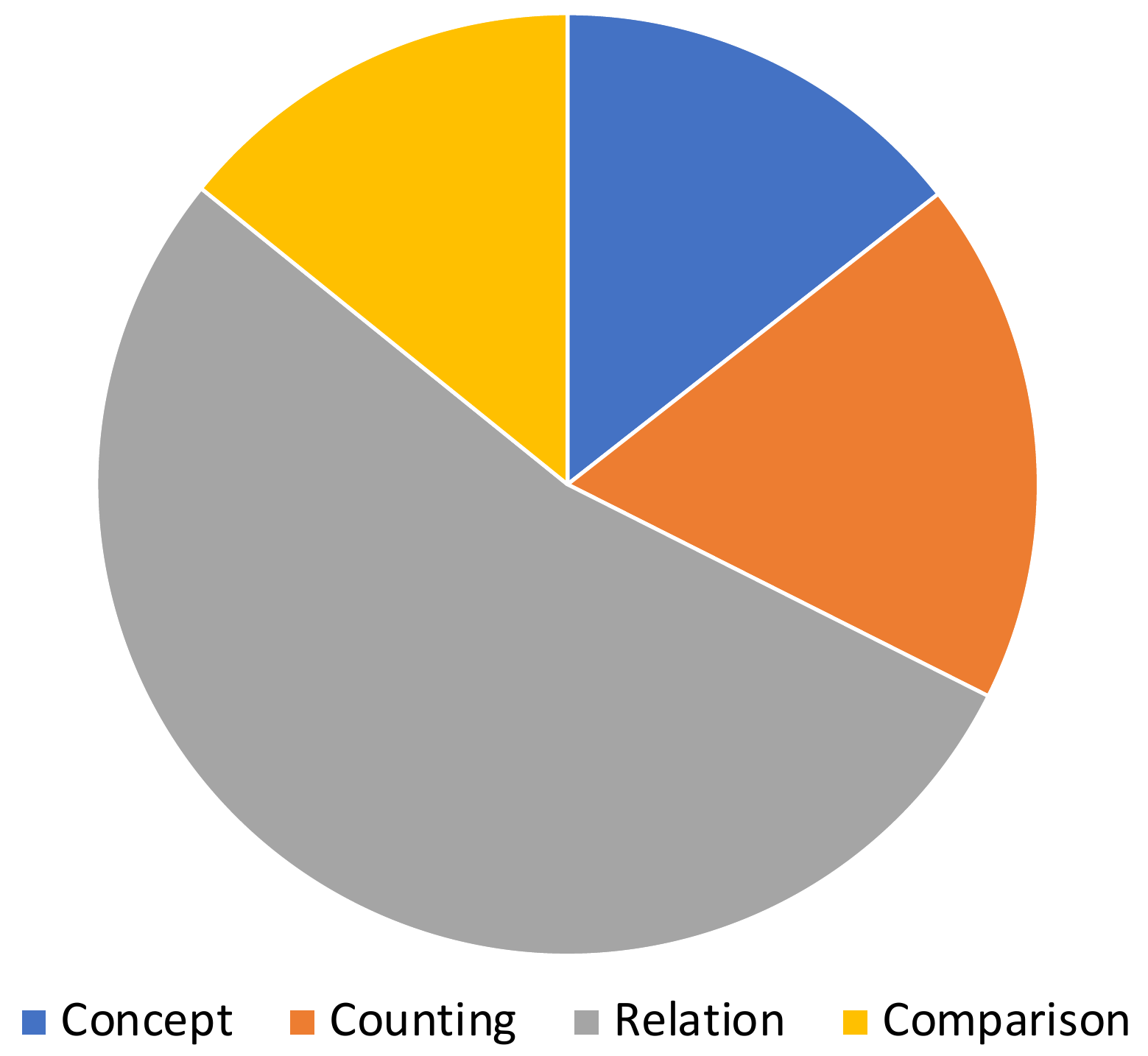}
\caption{Distribution of Question Type}
\label{fig:type}
\end{subfigure}\hspace{0.\textwidth}
\begin{subfigure}{.5\textwidth}
\includegraphics[width=\linewidth]{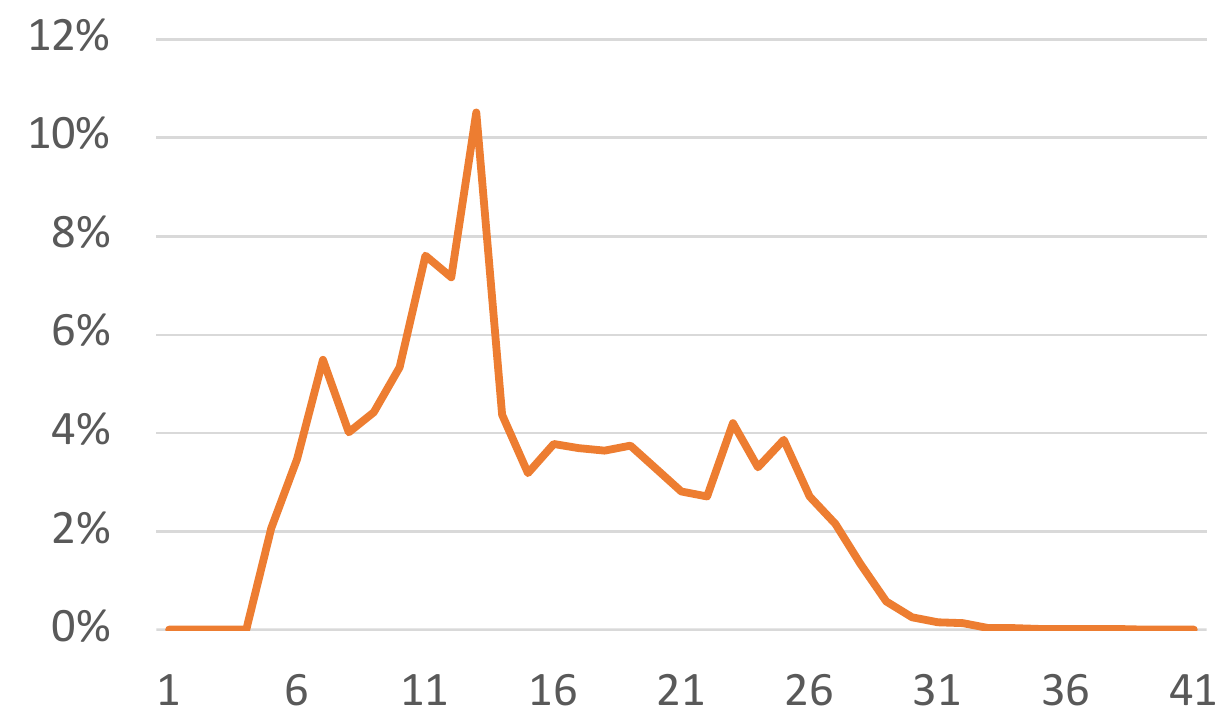}
\caption{Distribution of Question Word Length}
\label{fig:length}
\end{subfigure}
\caption{Dataset Statistics}
\label{fig:statistics}
\end{figure}

\subsection{More Dataset Examples}
In Figure \ref{fig:examples}, we show some more examples of our 3DMV-VQA dataset.
\begin{figure*}[htbp]
\centering
\includegraphics[width=0.8\textwidth]{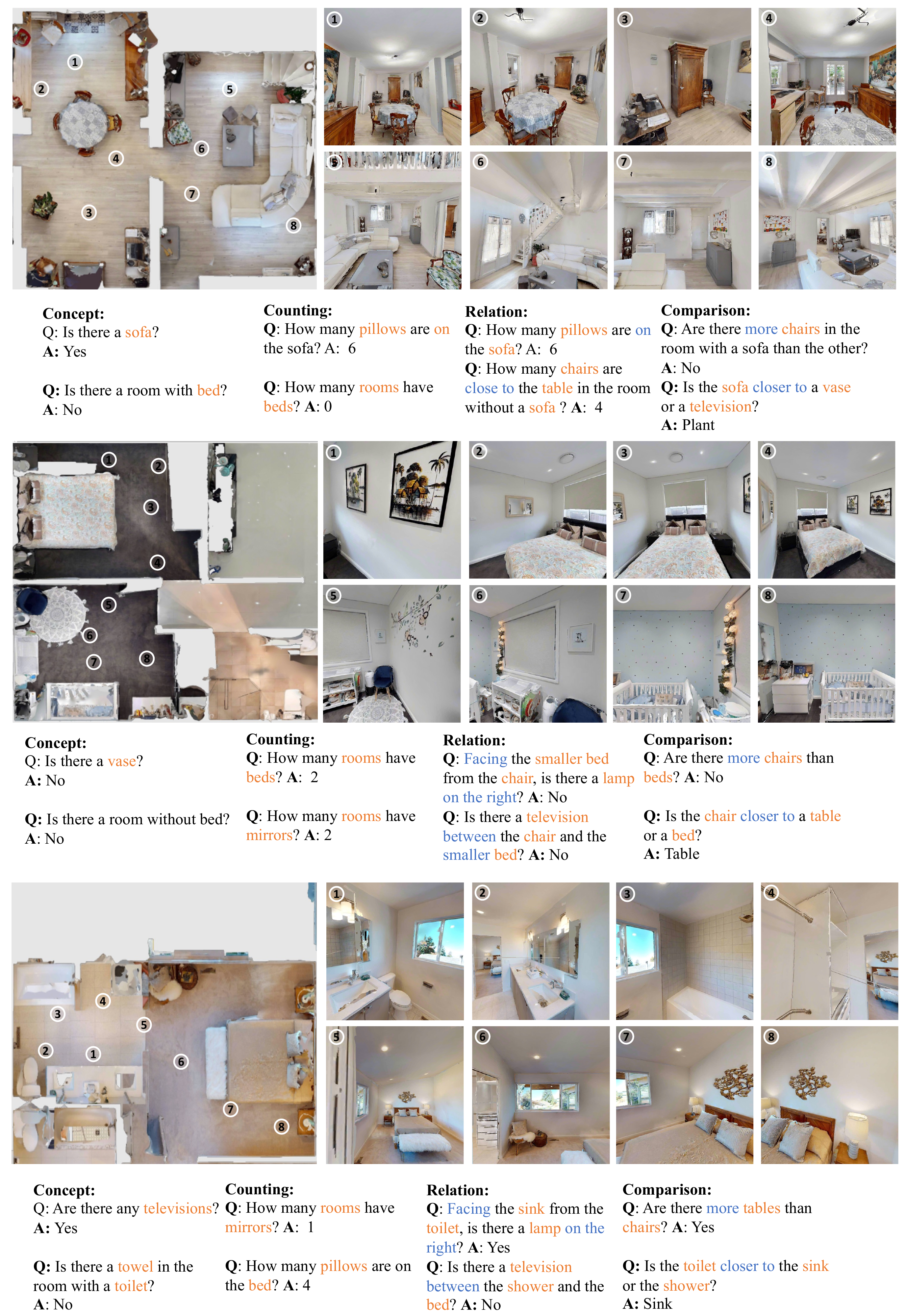}
\vspace{-3mm}
\caption{More Examples of 3DMV-VQA Dataset.}
\label{fig:examples}
\end{figure*}

\section{Implementation details}
\label{sec:details}
\subsection{Reasoning Operators}
\begin{itemize}
\item \textsc{Filter}
The \textsc{Filter} operator takes a voxel grid and a concept as input, and outputs the filtered voxel grid where points do not belong to the input have zero density values.

\item \textsc{Get\_Instance}
The \textsc{Get\_Instance} operator takes a voxel grid as input and outputs a list of voxel grids of different instances. We use DBSCAN, an unsupervised algorithm to assign the voxel grid points with densities greater than 0.5 into different instances. If the input is of the same semantic object, we directly use DBSCAN to get the instances. Else, we first get the voxel grids of different classes from semantic concept grounding, and then get the instances of all the instances of each semantic class. After that we integrate the instances of all semantic classes.

\item \textsc{Query}
The \textsc{Query} operator takes the voxel grid of an instance and returns the semantic concept of the instance.

\item \textsc{Count}
The \textsc{Count} operator takes a list of voxel grids as input and returns the length of the list.

\item \textsc{Exist}
The \textsc{Exist} operator takes a list of voxel grids as input and examines if the list is empty or not. If the list is empty, then the targeted concept doesn't exist.

\item \textsc{Get\_Room\_Instance}
The \textsc{Get\_Room\_Instance} operator takes a voxel grid as input and returns a list of new voxel grids of different room instances.
To get the instances of the rooms, we use the results of 3D semantic grounding and extract all walls. We then 
 segment the whole scene into rooms using the wall instances. 

\item \textsc{Filter\_Room}
The \textsc{Filter\_Room} operator takes a list of voxel grids of different room instances and returns another list of voxel grids where there are non-zero density values.

\item \textsc{Count\_Room}
The \textsc{Count\_Room} operator takes a list of voxel grids of different room instances and returns the length of the list of voxel grids where there are non-zero density values. It's a combination of the \textsc{Filter\_Room} operator and the \textsc{Count} operator.

\item \textsc{Exist\_Room}
The \textsc{Exist\_Room} operator takes a list of voxel grids of different room instances and returns whether the list of voxel grids where there are non-zero density values is empty or not. It's a combination of the \textsc{Filter\_Room} operator and the \textsc{Exist} operator.

\item \textsc{Relation}
The \textsc{Relation} operator takes a relation tuple (the voxel grids of two or three instances) and a relation, concatenate them and pass them into the relation module network of the specified relation, and outputs a score (which can be turned into True/False value according to whether the score is greater than 0.5 or not ) indicating whether the two/three instances have the relation or not.

\item \textsc{Filter\_Relation}
The \textsc{Filter\_Relation} operator takes two/three lists of voxel grids of different semantic classes and a specified relation. For all possible tuples chosen from the lists, each containing the two or three instances of different semantic classes, we pass the concatenated tuple into the relation module network and collects all True/False values. We filter out all tuples with true values and returns a list of the concatednated voxel grids.

\item \textsc{Exist\_Relation}
The \textsc{Exist\_Relation} operator takes two/three lists of voxel grids of different semantic classes and a specified relation. For all possible tuples chosen from the lists, each containing the two or three instances of different semantic classes, we pass the concatenated tuple into the relation module network and collects all True/False values. We examine whether there's a tuple with true value. It's a combination of the \textsc{Filter\_Relation} operator and the \textsc{Exist} operator.

\item \textsc{Count\_Relation}
The \textsc{Count\_Relation} operator takes two/three lists of voxel grids of different semantic classes and a specified relation. For all possible tuples chosen from the lists, each containing the two or three instances of different semantic classes, we pass the concatenated tuple into the relation module network and collects all True/False values. We count how many tuples with true values we have. It's a combination of the \textsc{Filter\_Relation} operator and the \textsc{Count} operator.

\item \textsc{Relation\_More}
The \textsc{Relation\_More} operator takes a specified relation in the comparison form (\textit{e.g.,} closer, more left), a first voxel grid, together with two second voxel grids for comparison. For each of the two second voxel grids, it's concatenated with the first voxel grid and input into the relation network. Then we output the voxel grid with the higher score value. 

\item \textsc{Relation\_Most}
The \textsc{Relation\_Most} operator takes a specified relation in the comparison form (\textit{e.g.,} closest, leftmost), a first voxel grid, together with a list of second voxel grids for comparison. For each of the second voxel grids, it's concatenated with the first voxel grid and input into the relation network. Then we output the voxel grid with the higher score value. 

\item \textsc{Larger\_than}
The \textsc{Larger\_than} operator takes as input two integers and returns whether the first integer is greater than the second.

\item \textsc{Smaller\_than}
The \textsc{Smaller\_than} operator takes as input two integers and returns whether the first integer is smaller than the second.

\end{itemize}

\subsection{Baselines}
\paragraph{CNN-LSTM, MAC \& MAC(V)} We use an ImageNet-pretrained ResNet-50 to extract 14 × 14 × 1024 feature maps for MAC and MAC(V). We use the 2048-dimensional feature
from the last pooling layer. 
\paragraph{3D-Feature + LSTM} We first use PCA to downgrade the feature size from 512 to 16. We then pass the the 3D-feature through three 3D-CNN (implemented by sparse convolutions) layers with intermediate size 64, to further downsample. We then concatenate this 3D feature with LSTM language feature and into them into a MLP to get the final answer. 

\paragraph{ALPRO} We end-to-end finetune ALPRO model on our dataset with the pre-trained checkpoint for 10 epochs. Our settings for finetuning is the same as the finetuning configurations for MSRVTT-QA in ALPRO paper. We take the 1500 answer candidates of MSRVTT-QA, and replace some irrelavent candidates with answers appear in our dataset. During inference, the model take as input image frames of shape $224\times224$, and output a set of classification probabilities of 1500. The predictions are obtained as the answer with the highest probability.  

\paragraph{LGCN} We take our QA-pairs as "FrameQA" task splited in LGCN paper on TGIF-QA dataset. For a QA-pair, we take image frames, each extracting 5 bounding boxes and their regional features of 1024-dim using MaskRCNN, and embed the question by word level and character level the same way as mentioned LGCN paper. Other settings for training and inference use same as FrameQA configurations.

\paragraph{NS-VQA} We first use CLIP-LSeg to get per-pixel semantic label to perform semantic concept grounding. After Filtering with certain concepts, all the pixels that do not belong to the concepts are set to white-transparent pixels. For counting problems, we also use DBSCAN which takes the pixel x-y values concatenated by their color values as input, and assign instance clusters to all the non-transparent pixels. For training of the relation network, we use pretrained ResNet-50 to get the 2D features of the images with filtered instances, concatenate them with language features, and then go through one MLP to output a score. We ``maxpool" the predictions in each image. For example, for concept questions that ask about whether there's a semantic class in the scene, we iterate through all images and get the prediction, if there's a prediction ``yes" in one of the images then the final prediction is also ``yes". For counting problems, we also iterate through and get integer predictions, and we get the largest prediction among all the images as our final prediction. For query problems, we iterate through all images and get the predictions which are concepts, we choose the concept which appears most frequently among the images.

\section{Experiments}
\label{sec:exp}

\subsection{Generalization Results}
\subsubsection{Generalization to Replica}
\begin{figure*}[htbp]
\centering
\includegraphics[width=\textwidth]{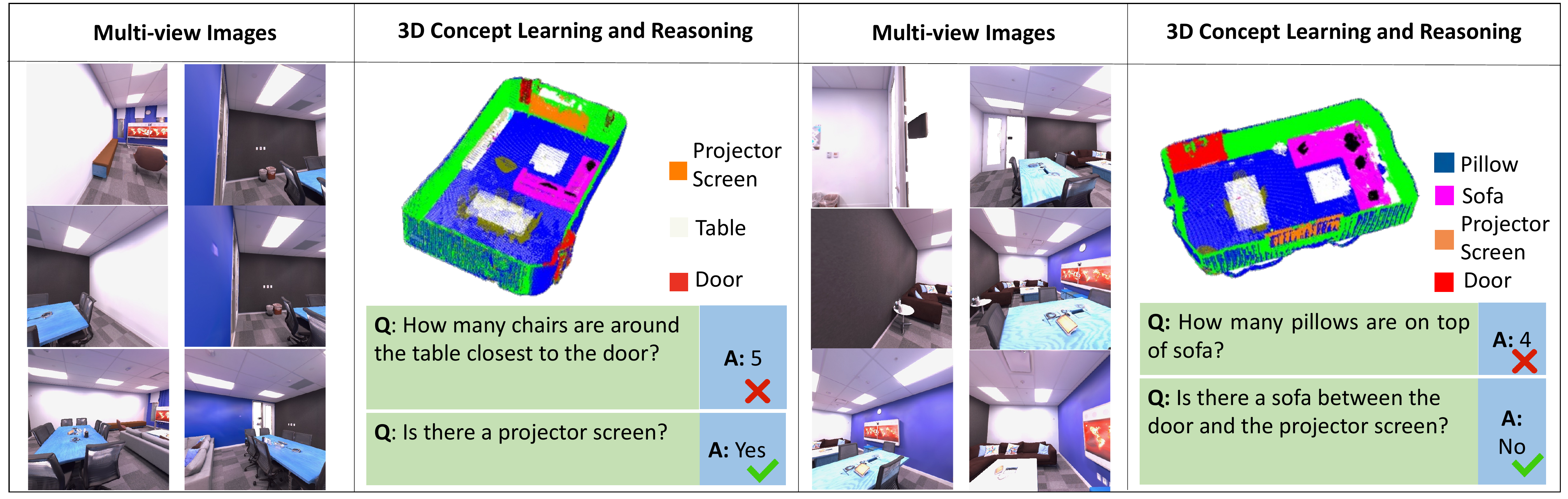}
\caption{Qualitative examples of generalizing to Replica dataset.}
\label{fig:replica}
\end{figure*}
\paragraph{Result Analysis} To further show that 3D-CLR trained on HM3D can be generalized to new reasoning datasets, we further collect a small visual question answering dataset on Replica \cite{Straub2019TheRD} with Habitat following the same data generation as HM3D, and with the same question type distribution as in Figure \ref{fig:type}. Table \ref{tab:replica} shows the results. We can see that 3D-CLR can maintain the performance on Replica as it performs on HM3D, which shows 3D-CLR's good generalization ability. Specifically, 3D-CLR and NS-VQA can maintain the performance on the conceptual questions, suggesting that CLIP-LSeg is able to perform semantic concept grounding on new concepts in the new dataset. Moreover, we see that the performance on counting problem is even better than that of HM3D, this is probably because Replica scenes are simpler and contains fewer details about tiny objects, thus making instance segmentation easier. As for relational problems, 3D-CLR can also maintain good results, showing that the relation networks training on HM3D can also be utilized on other datasets and further suggesting that the vocabulary of relations is limited yet general across all scenes, and can be learned from scratch.
\begin{table}[htbp]
	\begin{center}
	\begin{tabular}{lccccc}
	\toprule
      Methods   & Concept & Counting & Relation& Comparison \\ 

        MAC &55.7&16.4&40.9&58.8\\    
        MAC(V) &54.1&17.4&41.2&60.8\\
        NS-VQA &57.2&18.7&30.4&62.3\\ 

        \midrule
        3D-CLR  & \textbf{65.3} & \textbf{45.1} &\textbf{53.6}& \textbf{73.5}\\ 

    \bottomrule
	\end{tabular}
	\end{center}
 \vspace{-5mm}

	\caption{Question-answering accuracy of 3D visual reasoning baselines on different question types when generalizing to Replica dataset.}

	\label{tab:replica}
\end{table}
\paragraph{Qualitative Examples} In Figure \ref{fig:replica}, we show some qualitative examples of generalizing to Replica scenes. From the examples, we can come to several conclusions. First, 3D-CLR can perform zero-short semantic concept grounding on unseen concepts in HM3D, such as ``projector screen" and capture relations such as ``between". Second, it still performs poorly in counting questions. In the example on the left, it cannot count the instances of ``chairs" and on the right, it cannot count the instances of ``pillows". This is because the objects are ``sticked" to each other and are not separated in space. Therefore, DBSCAN cannot tell the objects apart.

\subsubsection{Generalization to Unseen Concepts}
To demonstrate 3D-CLR's zero-shot concept grounding ability and generalization ability, we generate some more question-answer pairs on unseen concepts. Recall that in the dataset section in the main paper, we would merge some concepts with similar concepts (\textit{e.g.,} ``stuffed animal" with ``toy"). To generate the datasets with unseen concepts, we use these merged concepts instead of the concepts in the proposed 3DMV-VQA dataset. We also append some unseen concepts manually to the new dataset. We assure that there are no overlapping words between the seen concepts and unseen concepts. Table \ref{tab:unseen} shows the generalization result. We can see that 3D-CLR and NS-VQA can still have good results on the conceptual problems, suggesting that they could perform zero-shot concept grounding. However, MAC and MAC(V) has very poor performances, much worse than the performances than HM3D. This suggests that the modular design and incorporation of CLIP-LSeg equips 3D-CLR with zero-shot generalization ability.

\begin{table}[htbp]
	\begin{center}
	\begin{tabular}{lccccc}
	\toprule
      Methods   & Concept & Counting & Relation& Comparison \\ 

        MAC &51.3&15.6&36.2&53.5\\    
        MAC(V) &51.4&16.1&38.5&54.2\\
        NS-VQA &58.6&19.2&29.7&58.1\\ 

        \midrule
        3D-CLR  & \textbf{63.4} & \textbf{37.7} &\textbf{55.1}& \textbf{68.9}\\ 

    \bottomrule
	\end{tabular}
	\end{center}
  \vspace{-4mm}

	\caption{Question-answering accuracy of 3D visual reasoning baselines on different question types when generalizing to unseen categories.}
 \vspace{-4mm}
	\label{tab:unseen}
\end{table}

\begin{figure*}[htbp]
\centering
\includegraphics[width=\textwidth]{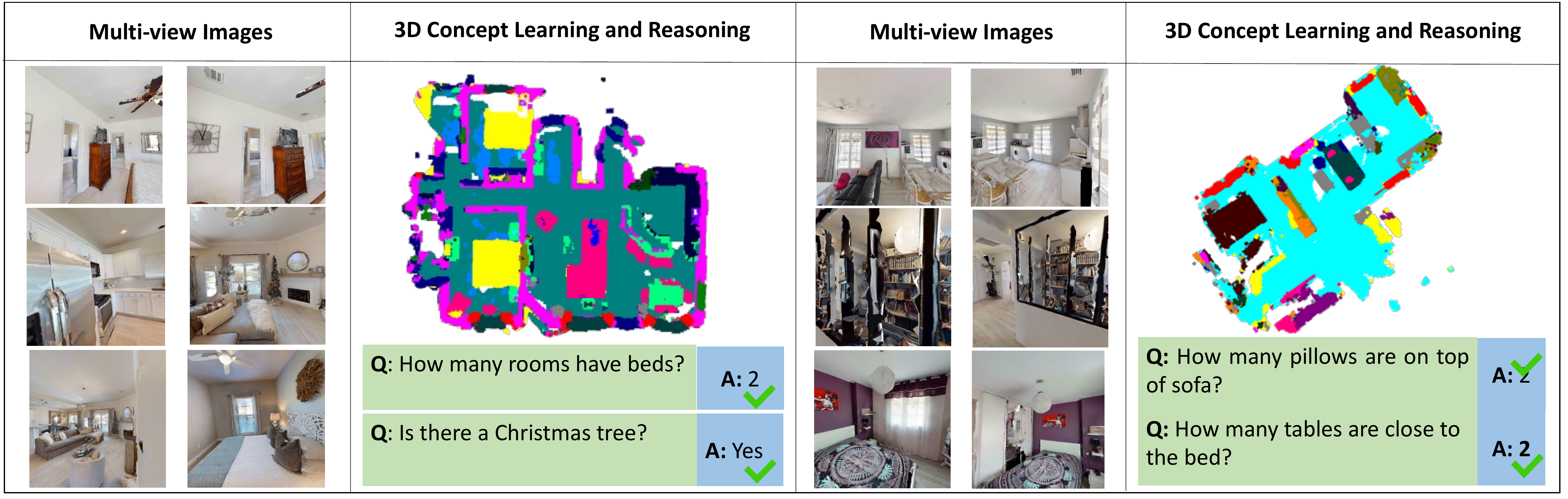}
\caption{More Qualitative Examples on 3DMV-VQA.}
\label{fig:qual_supp}
\end{figure*}

\subsection{More Qualitative Examples on 3DMV-VQA}
In Figure \ref{fig:qual_supp}, we show more qualitative examples on our 3DMV-VQA dataset. As we can see, 3D-CLR can generalize well to unseen concepts like ``Christmas tree", and can perform well on counting problems if the instances are well apart from each other.


 \fi

\end{document}